\def\year{2020}\relax
\documentclass[letterpaper]{article} 
\usepackage{aaai20}  
\usepackage{times}  
\usepackage{helvet} 
\usepackage{courier}  
\usepackage[hyphens]{url}  
\usepackage{graphicx} 
\usepackage{graphicx}  
\usepackage{amsmath}
\usepackage{amssymb} 
 \usepackage{bbm}
 \usepackage{array}
 \usepackage{multirow}
\usepackage{booktabs}  
\usepackage{threeparttable}
\usepackage{subcaption}
\usepackage{graphicx}
\title{Adversarial Attack on Deep Product Quantization Network for Image Retrieval}
\author{Yan Feng,\textsuperscript{\rm 1, \rm 2,}\thanks{The first two authors contribute equally to this work.} Bin Chen,\textsuperscript{\rm 1, \rm 2,}\footnotemark[1] Tao Dai,\textsuperscript{\rm 1, \rm 2,}\thanks{Corresponding author: Tao Dai} Shu-Tao Xia\textsuperscript{\rm 1, \rm 2}\\ 
\textsuperscript{\rm 1} Tsinghua Shenzhen International Graduate School, Tsinghua University, Shenzhen, China\\ 
\textsuperscript{\rm 2} PCL Research Center of Networks and Communications, Peng Cheng Laboratory, Shenzhen, China\\
y-feng18@mails.tsinghua.edu.cn, cb17@mails.tsinghua.edu.cn, daitao.edu@gmail.com, xiast@sz.tsinghua.edu.cn
}
 
\begin{document}

\maketitle

\begin{abstract}
Deep product quantization   network (DPQN)   has recently received much attention in fast image retrieval tasks due to its efficiency of encoding high-dimensional visual features especially when dealing with large-scale datasets. 
Recent studies show that deep neural networks (DNNs) are vulnerable to input with small and maliciously designed perturbations (a.k.a., adversarial examples).
This phenomenon raises the concern of security issues for DPQN in the testing/deploying stage as well. However, little effort has been devoted to investigating how adversarial examples affect DPQN. 
To this end, we propose product quantization adversarial generation (PQ-AG), a simple yet effective method to generate adversarial examples for product quantization based retrieval systems. 
 PQ-AG aims to generate imperceptible adversarial perturbations for query images to form adversarial queries, whose nearest neighbors from a targeted product quantizaiton model are not semantically related to those from the original queries.
Extensive experiments show that our PQ-AQ successfully creates adversarial examples to mislead targeted product quantization retrieval models. Besides, we found
that our PQ-AG significantly degrades retrieval performance in both white-box and black-box settings.

\end{abstract}

\section{Introduction}

The massive application of large scale and high dimensional image data has attracted great attention to the study of efficient image retrieval methods. In order to strike the balance between storage and precision, approximate nearest neighbor (ANN) based retrieval methods \cite{datar2004locality,gionis1999similarity,muja2009fast,MITbook} have been widely studied. Among these methods, product quantization (PQ) based retrieval \cite{pq,OPQ} 
becomes popular due to the efficiency and effectiveness of PQ for coding high-dimensional visual features. Recently proposed  deep product quantization network (DPQN) based methods \cite{CaoL0ZW16,Klein_2019_CVPR,eccv/YuYFJ18,liu2019deep} 
combine the representative power of deep  neural  networks (DNNs) with PQ, and achieve impressive performance and efficiency.
\begin{figure}[tb]
\begin{center}
\includegraphics[width=0.95\columnwidth]{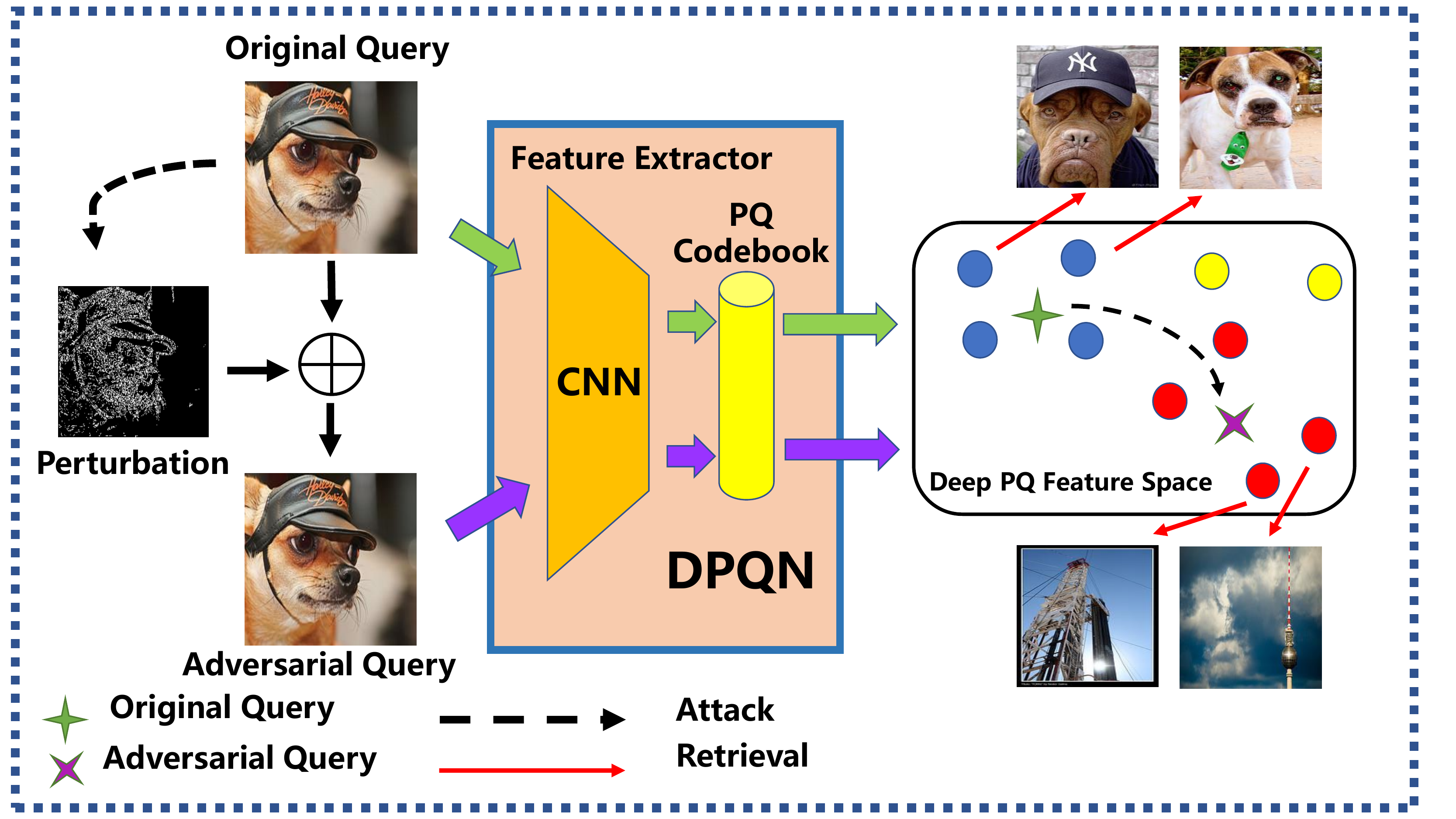}
\end{center}
  \caption{An example of adversarial attack against DPQN. Although the added perturbation is almost invisible, it successfully causes the query to shift a lot in deep PQ feature space, and fools the DPQN to retrieve semantically irrelevant images.}
\label{intro}
\end{figure}

On the other hand, recent studies show that DNNs are vulnerable to input images with carefully designed yet imperceptible perturbations, i.e., adversarial examples, which can cause DNN classifier to produce confidently wrong predictions \cite{corr/SzegedyZSBEGF13}. To date, much attention has been drawn to crafting adversarial examples in image classification task, while little effort has been devoted to investigating how adversarial examples affect  DPQN for image retrieval.  This raises the concern of security issues of DPQN in practice. As shown in Fig.~\ref{intro}, DPQN based image retrieval system is successfully misled by an adversarial query image with imperceptible perturbations.
 Note that  the well-known adversarial attacks for image classification can be interpreted as maximizing the classification error via tuning the input image according to the direction of gradients. 
However, attacking DPQN retrieval systems can be much more challenging than the classification task for two main reasons: 1) in an image retrieval task,  a query image is searched through a database on visual features in the absence of class labels; 2) the optimization of adversarial attacks generally relies on back-propagation that requires differentiable operations, while DPQN consists of  product quantization operation, which is  indifferentiable.

Motivated by the above observations,  we propose  product quantization adversarial generation (PQ-AG), a simple yet  effective method to craft adversarial examples for deep product quantization models. Given a clean query image and a targeted product quantization retrieval model, our proposed PQ-AG method can generate adversarial queries with small and well-crafted perturbations    to fool the targeted model to retrieve semantically irrelevant images from the database.
Specifically, we first formulate the objective as a feature separation problem, which pushes the features of adversarial query away from those of the original neighbors.  Furthermore, to deal with the indifferentiable issue of PQ, we alternatively propose to maximize the difference of the codewords assignment probability distribution between the adversarial query and the original one, where the distributions can be estimated based on the similarity between features and codewords.

Our main contributions  can be  summarized as follows:
\begin{itemize}
\item We propose a simple yet effective product quantization adversarial generation (PQ-AG) method to mislead the DPQN based image retrieval systems. To the best of our knowledge, this is the first attempt of adversarial attack design targeting on the DPQN based image retrieval system. 
\item In order to tackle the indifferentiable problem of product quantization, we  propose a novel strategy to design an adversarial query by perturbing the probability distribution of codewords assignments for a clean query.
\item We evaluate PQ-AG on a variety of datasets and model structures. Experimental results demonstrate the effectiveness and transferability of our method in both white-box and black-box settings. 
\end{itemize}

\begin{figure*}[ht]
\begin{center}
\includegraphics[width=0.95\textwidth]{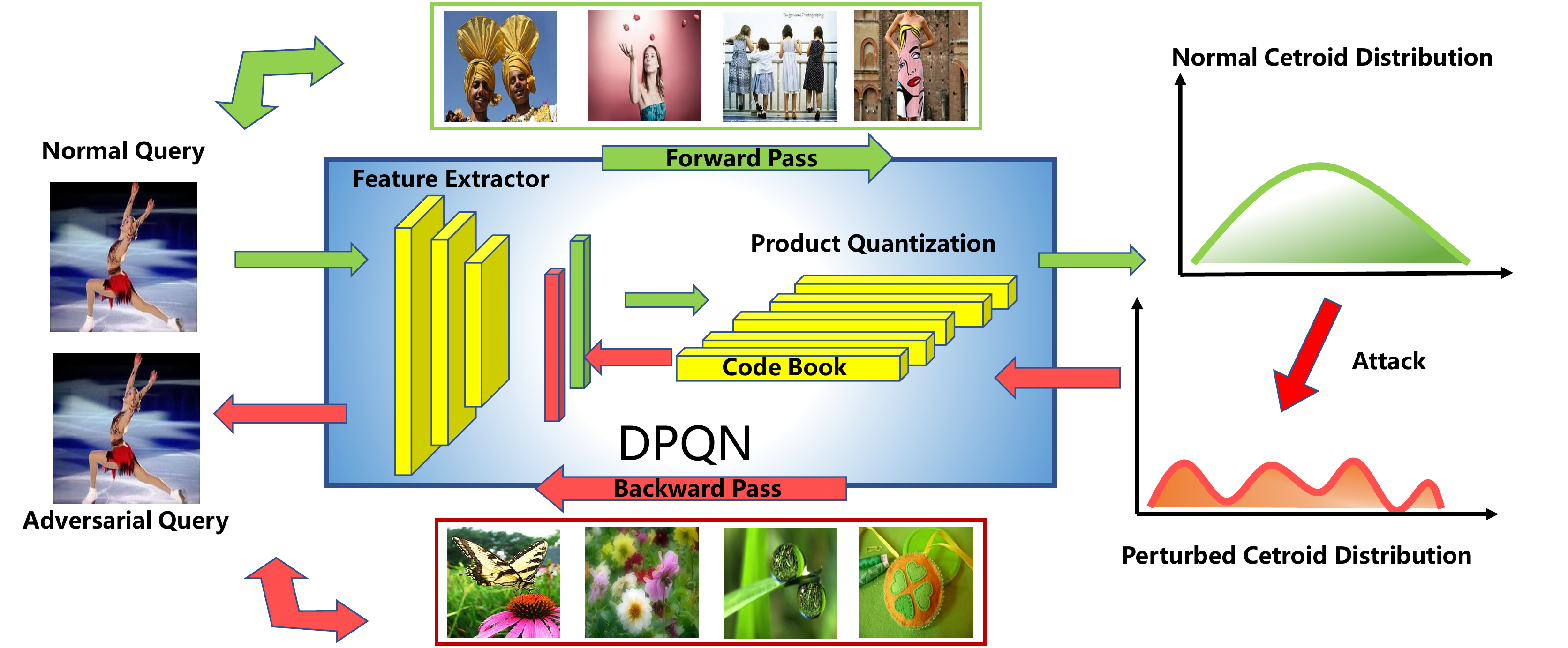}
\end{center}
  \caption{The overall pipeline of the proposed PQ-AG method against DPQN based retrieval systems. A query is first fed into the CNN model, then the normal centroid distribution is computed based on the cosine similarity between the deep feature and centroids of the codebook. The goal of PQ-AG is to perturb the distribution, thus disrupting the subsequent retrieval operation. Gradients of the perturbation will be back-propagated to update the adversarial query. }
\label{pipeline}
\end{figure*}
\section{Related Work}
We give the formal definition of PQ and briefly review recent advances of PQ based  retrieval methods applied in lager scale and high dimension datasets. Then we also provide a brief introduction to some related works on adversarial attack.
\subsection{PQ based Image retrieval systems}
To efficiently conduct the approximate nearest neighbor (ANN) search for image retrieval in large scale databases, existing mainstream approaches can be divided into two two categories: 1) hashing based methods 2) PQ based methods. Hashing method aims at learning a hash function, which maps the original data points to a binary code. 
The main advantage of hashing methods is  the fast-retrieval speed due to the computation by hamming distance and low storage overhead. Unlike hashing methods that can only produce a few distinct distances, PQ based methods can better describe the distance between data points, whose goal is to employ K-means individually in each subspace of original data points or their transformed ones. Next, we will briefly describe the core of  PQ based image retrieval methods.

Formally, suppose that $\mathcal{D}\subseteq \mathbb{R}^{D}$ is a database containing $N$ data points, a given vector ${\bf x} \in \mathcal{D}$ is first partitioned into $M$ sub-vectors with equal dimension $D/M$, i.e. ${\bf x}=({\bf x_1}, {\bf x_2}, ..., {\bf x_M})$. Then each subvector ${\bf x_m}$, $m\in[M]\triangleq \{1, 2, \dots, M\}$, is quantized into the $K$ centroids within a sub-codebook $\mathcal{C}_m=\{{\bf c_{m1},\dots, {\bf c_{mK}}}\}$ in each subspace uniquely as
\begin{equation}\label{hard assignment}
q({\bf x_m}) = \arg\min \limits_{k} ||{\bf x_m} - {\bf c_{mk}}||_2^2,
\end{equation}
where ${\bf c_{mk}}$ is the $k$-th centroid in the $m$-th sub-codebook. Therefore, every data point ${\bf x}$ can be approximately represented by its corresponding centroids, i.e.,  $${\bf x}\approx q({\bf x})= (q({\bf x_1}), ..., q({\bf x_M})).$$
Note that PQ can generate a Cartesian product codebook $\mathcal{C}=\mathcal{C}_1\times\mathcal{C}_2\times\dots\mathcal{C}_M$ with an exponential size. The cost of the storage is low with $\mathcal{O}(DK)$ and the retrieval speed is fast due to the use of look-up tables. 

In the retrieval stage, PQ-based methods conduct approximated nearest neighbor (ANN) based on similarity measure between two images in feature space.
There are two main types of   similarity measures, i.e., symmetric distance computation (SDC) and asymmetric distance computation (ADC), to approximate Euclidean distance computation between the query ${\bf y}$ and the vector ${\bf x}$ in the database,  which is defined  as follows:
\begin{equation*}
\!\!\! d_{\text{SDC}}({\bf y}, {\bf x}) \approx d(q({\bf y}), q({\bf x}))= \sqrt{\sum_{m=1}^{M}d(q({\bf y_m}), q({\bf x_m}))^2},
\end{equation*}
\begin{equation*}
\!\!\! d_{\text{ADC}}({\bf y}, {\bf x}) \approx d({\bf y}, q({\bf x}))= \sqrt{\sum_{m=1}^{M}d({\bf y_m}, q({\bf x_m}))^2},
\end{equation*}
where $d(\cdot,\cdot)$ denotes the Euclidean distance.

Recently, many PQ based methods  \cite{pq,OPQ,wang2014optimized,kalantidis2014locally,AQ,SQ,yu2017bilinear,wu2017multiscale,li2017distribution,CQ} have been developed for fast image retrieval.
To obtain better retrieval performance, deep product quantization network (DPQN) based methods \cite{CaoL0ZW16,Klein_2019_CVPR,liu2019deep,eccv/YuYFJ18} have been recently proposed by  combining PQ and deep convolutional neural network. \citeauthor{CaoL0ZW16}~\shortcite{CaoL0ZW16} gave a step forward in this direction and proposed the deep quantization network (DQN) to capture good hashing based image representations under a similarity-preserving and a product quantization loss. Later, deep product quantization (DPQ) \cite{Klein_2019_CVPR} was developed by cluster optimization of PQ with the supervised signal and the parameters update of the fully-connected layer simultaneously.
\citeauthor{liu2019deep}~\shortcite{liu2019deep} proposed the deep triplet quantization (DTQ) to learn deep quantization models by introducing a novel similarity triplet loss in the training process.
To better incorporate PQ in a neural network, product quantization network (PQN) \cite{eccv/YuYFJ18} adds a soft product quantization layer to the neural network and obtains a clever codewords assignment according to the similarity between the features, which mitigates the over-fitting problem and consistently outperforms other methods. Despite of  many  DPQN based retrieval methods, little effort has been devoted to the security issue of DPQN for image retrieval, which is the main reason why we investigate how adversarial examples affect DPQN.

\subsection{Adversarial Examples}
Recent studies show that deep neural networks (DNNs) are vulnerable to input with small and maliciously designed perturbations, i.e., adversarial examples. Depending on the access permission to  the information of the neural network, including the neural structure and parameters, adversarial attacks can be roughly divided into two types: white-box \cite{corr/SzegedyZSBEGF13,corr/GoodfellowSS14} and black-box attacks \cite{sp/Carlini017,corr/GoodfellowSS14,eccv/BhagojiHLS18}. White-box attacks can obtain better performance, since they can easily access to the gradient information of DNNs. Among them, 
\citeauthor{corr/SzegedyZSBEGF13}~\shortcite{corr/SzegedyZSBEGF13} recently showed that some targeted neural networks can be easily fooled by crafted adversarial examples. Other attack algorithms like fast gradient sign method (FGSM) based methods \cite{corr/GoodfellowSS14,iclr/MadryMSTV18} also obtain remarkable performance. Instead, black-box attacks \cite{sp/Carlini017,corr/GoodfellowSS14,cvpr/Moosavi-Dezfooli17,eccv/BhagojiHLS18} are more challenging, since they  cannot access  to the information of DNNs.
 Besides these attacks on image classification task, some other methods focus on the adversarial attack on other tasks, including image captioning \cite{corr/abs-1712-02051} and semantic segmentation \cite{iccv/XieWZZXY17}. 
 
 However, little effort has been devoted to  the image retrieval task. To the best of our knowledge, two related works, i.e.,  hash adversary generation (HAG) \cite{Yang2018Adversarial} and UAA-GAN \cite{Zhao2019Gan}, both focus on adversarial attack on image retrieval. Nevertheless, there exist two main  differences compared with our method.
1) Our work is a first attempt to consider the adversary generation for product quantization based image retrieval, while HAG and UAA-GAN focus on hashing based adversarial perturbation generation and generative network design for conventional retrieval systems, respectively; 2) our work focuses on designing a simple yet effective attack method  in product quantization feature space, which is more challenging, while HAG and UAA-GAN concentrate on attacking in deep feature space.

\section{Product Quantization Adversarial  Generation}

In this section, we would demonstrate how product quantization adversarial  generation (PQ-AG) algorithm generates adversarial queries to fool the DPQN based retrieval method into retrieving semantically irrelevant images. As shown in Fig.~\ref{pipeline},  our PQ-AG can generate adversarial query by perturbing the centroid distribution in product quantization space to attack DPQN. 
\subsection{Formulation of the Overall Objectives}
In an image retrieval task, adversarial query generation algorithms for a targeted retrieval system usually consider the following problem:
for a given query ${\bf y}$, the goal of adversarial query generation is to generate its corresponding adversarial example ${\bf \hat{y}}$, whose retrieval results are semantically irrelevant to ${\bf y}$.  Denote $F(\cdot,\cdot)$ as a feature-dependent transformation functions whose output contains the semantic information of the original feature. We further denote $d(\cdot,\cdot)$ as a metric to measure the similarity between $F({\bf y})$ and $F(\hat{{\bf y}})$. Therefore, the core idea of generating adversarial queries  is to expand the distance as defined by metric  $d$ between the semantic features $F(\hat{{\bf y}})$ and $F({\bf y})$ for  adversarial and clean queries. 

To this end, we can formulate the adversarial query generation problem as follows:
\begin{equation}\label{general}
\begin{split}
   \min \limits_{\hat{{\bf y}}} &\,\,\, \mathcal{L}({\bf y}, \hat{{\bf y}}) = -d\left(F({\bf y}), F(\hat{{\bf y}})\right)\\
\text{s.t.}&\,\,\,\quad ||{\bf y}-\hat{{\bf y}}||\leq \eta,
\end{split}
\end{equation}
 where $d(\cdot,\cdot)$ is a similarity measure function, e.g. Euclidean distance, $||\cdot||$ can be any norm specified by the user, e.g., $\ell_0$, $\ell_1$, or $\ell_{\infty}$-norm,  and $\eta>0$ denotes the magnitude of the attack. 
 
For example, if we  set $F(\cdot, \cdot)$ to be a CNN feature extractor $F_{\text{CNN}}(\centerdot)$ and $d(\cdot, \cdot)$ to be Euclidean distance, then the objective function (\ref{general}) can be formulated as:
\begin{equation}\label{basic}
\begin{split}
  \min \limits_{\hat{{\bf y}}} &\,\,\, \mathcal{L}({\bf y}, \hat{{\bf y}}) = -|| F_{\text{CNN}}({\bf y}), F_{\text{CNN}}(\hat{{\bf y}})||_2^2\\
\text{s.t. }&\,\,\, ||{\bf y}-\hat{{\bf y}}||\leq \eta.
\end{split}
\end{equation}
In practise, (\ref{basic}) represents the attack on a general deep feature based retrieval system. Therefore, we adopt (\ref{basic}) as the baseline method in our experiments.
 
In order to design a more successful attack against DPQN, we need have a closer look at its retrieval process. As shown in Fig.~\ref{pipeline},  DPQN  mainly consists of a feature extractor and a PQ codebook. 
During retrieval process, the input query is first fed into the feature extractor, which is usually a pre-trained convolutional neural network (CNN). Afterwards, the features are quantized as the closest centroids from the codebook, and the quantized feature can be written as:
 \begin{equation}
     F({\bf y})=q\left(F_{\text{CNN}}({\bf y})\right),
 \end{equation}
 where $q(\cdot)$ denotes the product quantization operation as (\ref{hard assignment}), which will be also referred to as hard centroid assignment operation in the subsequent parts.  
The retrieval results can then be determined as the nearest neighbors of the product-quantized feature from the database. 

It is clear that product quantization plays a key role in the above DPQN based retrieval process. Therefore, attacking on the quantized feature could be an effective way for adversarial query generation.

The specific form of (\ref{general}) against DPQN  can thus be formulated as:
\begin{equation}\label{indifferentiable}
\begin{split}
   \min \limits_{\hat{{\bf y}}} &\,\, \mathcal{L}({\bf y}, \hat{{\bf y}}) =-|| q\left(F_{\text{CNN}}({\bf y})\right)-q\left(F_{\text{CNN}}({\bf \hat{y}})\right)||_2^2\\
&\text{s.t.}\,\,\quad ||{\bf y}-\hat{{\bf y}}||\leq \eta.
\end{split}
\end{equation}

The next step is to optimize the objective in (\ref{indifferentiable}).
Since the function $q(\cdot)$ in (\ref{indifferentiable}) involves the indifferentiable operation, i.e., hard codewords assignment,  it is infeasible to directly optimize (\ref{indifferentiable}) based on  back propagation. To solve this problem, we alternatively propose the soft codewords assignment, whose details are shown in the following sections.

\subsection{PQ based Attack via Centroid Distribution Perturbation}
The proposed PG-AG seeks to generate adversarial examples in PQ feature space. From the above analysis, the main challenge of our PQ-AG lies in the indifferentiable issue of PQ.   
To solve this issue, we alternatively estimate the probability distribution of codewords assignment of a clean query based on the hard version (\ref{hard assignment}), i.e., one-hot encoding.

\subsubsection{Attack on the Peak of the Centroid Distribution (Type I Attack)}
Denotes ${\bf z}$ as the $l_2$-normalized feature extracted from  a pre-trained CNN for ${\bf y}$. Then the sub-vectors of ${\bf z}$ is assigned to the optimal centroids of sub-codebook $\{\bf c_{mk}\}_{m=1,k=1}^{M,K}$, and the indices $(b_1, b_2, \dots, b_m)$ of ${\bf z}=({\bf z_1}, {\bf z_2}, ..., {\bf z_M})$ can be specifically computed by
\begin{equation}\label{hard}
b_m=\arg\max_k\,\,\langle {\bf z_m, c_{mk}}\rangle, \,\, m\in[M].
\end{equation}
In order to solve the gradient-vanishing problem,  we adopt the soft centroid assignment first proposed in \cite{eccv/YuYFJ18}. In detail, we calculate the soft probability distribution ${\bf \hat{p}_m}=(\hat{p}_{m1},\hat{p}_{m2}, \dots, \hat{p}_{mK})$ of assigning the subvector ${\bf \hat{z}_m}$ of ${\bf \hat{z}}=F_{\text{CNN}}({\bf \hat{y}})$ to the centroids $\{\bf c_{mk}\}_{k=1}^{K}$ based on the respective cosine similarities as follows.

\begin{equation}\label{softassignment}
\hat{p}_{mk} = \frac{e^{\langle {\bf \hat{z}_m, c_{mk}\rangle}}}{\sum_{k^{'}} e^{\langle {\bf \hat{z}_m, c_{mk'}\rangle}}},\,\,k\in[K].
\end{equation}
It is easy to obtain the ground-truth probability distribution ${\bf p_m}$, i.e., one-hot encoding of the clean query ${\bf y}$ according to  (\ref{hard}). If we take the cross-entropy loss as our metric $d$ to measure the similarity between ${\bf \hat{p}_m}$ and ${\bf p_m}$, then the objective function (\ref{general}) can  be specifically formulated as:
\begin{equation}\label{hardsoftattack}
\begin{split}
   \min \limits_{\hat{{\bf y}}} &\,\,\, \mathcal{L}({\bf y}, \hat{{\bf y}}) =\sum_{m} -d({\bf \hat{p}_m}, {\bf p_m})\\
  &= \sum_{m} \sum_{k} \log (\hat{p}_{mk}) \mathbbm{1}(k = b_m) \\
  &= \sum_{m} \sum_{k} \log \left(\frac{e^{\langle {\bf \hat{z}_m, c_{mk}\rangle}}}{\sum_{k^{'}} e^{\langle {\bf \hat{z}_m, c_{mk'}\rangle}}}\right) \mathbbm{1}(k = b_m),\\
\text{s.t.}&\,\,\,\,\,\, ||{\bf y}-\hat{{\bf y}}||\leq \eta,
\end{split}
\end{equation}
where $\mathbbm{1}(\cdot)$ is an indicator function.

 Note that the above objective (\ref{hardsoftattack}) bears a resemblence to the loss commonly used in untargeted attack for classification tasks, where the $\hat{p}_{mk}$ corresponds to the probability of classifying the adversarial example to the $k_{th}$ class and $b_m$ corresponds to the true label. 
 As shown in Fig. \ref{eh2}, this kind of attack effectively decrease the probability of selecting the peak centroid of original distribution.
\subsubsection{Attack on the Overall Centroid Distribution (Type II Attack)}
Type I attack, however, may still have some limitations due to the following two key observations for symmetric and asymmetric retrieval attacks.

\begin{figure}[thb]
\centering
\begin{subfigure}{.47\linewidth}
  \centering
  \includegraphics[width=\columnwidth]{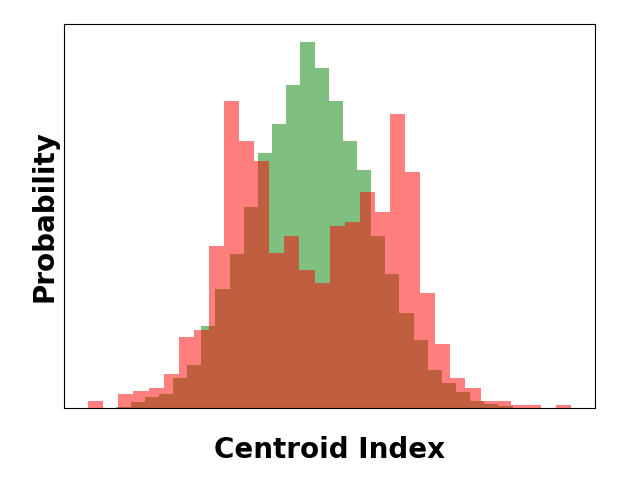}
 \caption{Type I Attack} 
  \label{eh2}
\end{subfigure}
\begin{subfigure}{.47\linewidth}
  \centering
  \includegraphics[width=\columnwidth]{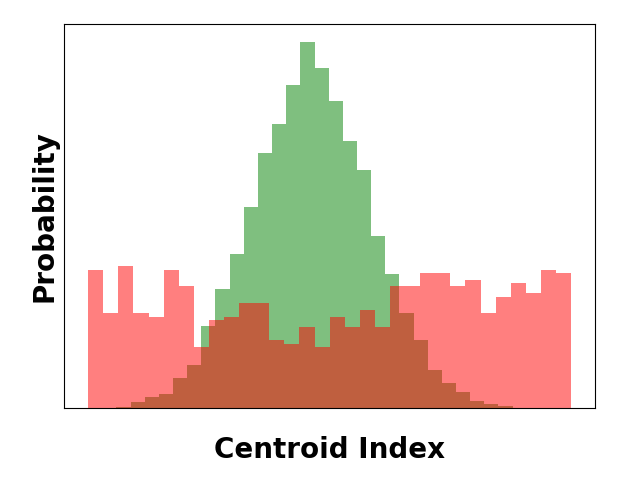}
 \caption{Type II Attack} 
  \label{eh3}
\end{subfigure}
 \caption{Centroid distribution under Type I and Type II attack. Normal distrbution is plotted with green color while the perturbed distribution with red color. Compared with Type I attack, Type II attack disrupts the overall distribution instead of simply focusing on the peak.}
\label{distribution}
\end{figure}

\begin{itemize}
\item In the phase of the symmetric retrieval, both the feature of query and database images will be quantized. Fig.~\ref{distribution} shows
that attacking on the peak will merely push the original centroid distribution towards a multi-peak distribution, of which the peaks correspond to the originally high-probability centroids. Moreover, those centroids with high probabilities, i.e., centroids with small angular difference, are very similar. 
Therefore, for a symmetric retrieval system, optimizing  (\ref{hardsoftattack}) may simply replace the hard-quantized centroids of a clean query ${\bf y}$ with slightly different ones, thus unable to effectively increasing the quantization error or attack effect.
\item In the phase of the asymmetric retrieval, only the feature of database images will be quantized. Therefore the similarity score or relevance between a query feature ${\bf z}=({\bf z_1}, {\bf z_2}, ..., {\bf z_M})$ and a database feature ${\bf x}=({\bf x_1}, {\bf x_2}, ..., {\bf x_M})$ is defined as 
$$s({\bf z}, {\bf x})=\sum_{m=1}^{M}\langle \bf z_m, c_{mb_m} \rangle,$$
where $b_m=\arg\max_k\,\,\langle {\bf x_m, c_{mk}}\rangle, \,\, m\in[M]$. Note that the scores
are clearly determined by the inner product pairs $\{\langle {\bf z_m, c_{mk} \rangle}\mid m\in[M], k\in[K]\}$ and implicitly determined by the distribution  $\{p_{mk}\}_{m=1,k=1}^{M, K}$ computed by  (\ref{softassignment}).
Consequently, it's crucial to distort the overall distribution instead of focusing purely on attacking the peak of the distribution of a clean query ${\bf y}$ as in (\ref{hardsoftattack}).
\end{itemize}

The above two observations motive us to pay more attention to perturbing the overall soft-quantized distributions  $\{p_{mk}\}_{m=1,k=1}^{M, K}$ of a clean query ${\bf y}$. To this end, we replace the one-hot distribution in  (\ref{hardsoftattack}) with the soft distribution $\{p_{mk}\}_{m=1,k=1}^{M, K}$ as (\ref{softassignment}) to learn an adversarial query ${\bf \hat{y}}$. 
Formally, we reformulate the objective  (\ref{general})  as follows:
\begin{equation}\label{softattack}
\begin{split}
   \min \limits_{\hat{{\bf y}}} &\,\,\, \mathcal{L}({\bf y}, \hat{{\bf y}})=\sum_{m} \sum_{k}p_{mk} \log (\hat{p}_{mk}) ,\\
\text{s.t.}&\,\,\, ||{\bf y}-\hat{{\bf y}}||\leq \eta.
\end{split}
\end{equation}

Note that optimizing (\ref{softattack}) is equivalent to maximize the KL divergence between the real distributions $\{p_{mk}\}_{m=1,k=1}^{M, K}$ and adversarial distributions $\{\hat{p}_{mk}\}_{m=1,k=1}^{M, K}$. In this way,  we can completely push the feature away from those centroids with high probabilities. Especially for the asymmetric attack, we have expanded the distance between the adversarial distribution $\{\hat{p}_{mk}\}_{m=1,k=1}^{M, K}$ and the original one, thus differentiating the similarity scores for ${\bf \hat{y}}$ and ${\bf y}$. As illustrated in Fig.~\ref{eh3},  Type II Attack leads to a more disordered distribution, which is harder to defend.
More results about our Type I and II Attacks are shown in the experiment section.

\section{Experiments}

In this section, we evaluate our proposed product  quantization  adversarial generation (PQ-AG) algorithm on two public benchmark datasets: {\bf CIFAR-10} and {\bf NUS-WIDE}. We first introduce the datasets, evaluation metric and implementation details. Then we  present and analyze the experimental results in detail.

\subsection{Datasets and Metrics}
\textbf{CIFAR-10} \cite{krizhevsky2009learning} contains 60000 color images of size 32 $\times$ 32 divided into 10 classes, each of which contains 6000 images. Following conventional configuration \cite{eccv/YuYFJ18}, \cite{CaoL0ZW16}, the training of targeted product quantization retrieval model is conducted on the training set containing 50000 images, while the test set is randomly divided into 9000 images as a  database and 1000 images as queries.

\noindent \textbf{NUS-WIDE} \cite{civr/ChuaTHLLZ09} is a large scale dataset for multilabel classification tasks. NUS-WIDE consists of 269648 images and each image is assigned with one or multiple labels related to 81 concepts. We select the subset of 186577 images associated with the 10 most popular concepts. The training set and test set are separated as default setting in \cite{civr/ChuaTHLLZ09}. For the generation of adversarial queries, 1000 images are randomly selected from the test set as query images, while the remaining images are used as database images.

\noindent \textbf{Evaluation Metrics}
We adopt the standard metric in image retrieval tasks, mean Average Precision (mAP), to evaluate the effectiveness of our methods. For the fairness of comparison, we evaluate the retrieval performance in both SDC and ADC settings, where SDC retrieval systems symmetrically quantize both query and database images and ADC systems quantize query images only. 

\begin{table*}[ht]
\begin{center}
\resizebox{0.95\textwidth}{!}{
\begin{tabular}{c<{\centering}  c<{\centering} c<{\centering} c<{\centering} c<{\centering} c<{\centering} c<{\centering} c<{\centering} c<{\centering} c<{\centering} c<{\centering} c<{\centering} c<{\centering} c<{\centering}}
\toprule
\multicolumn{2}{c}{}  & \multicolumn{6}{c}{ CIFAR-10} & \multicolumn{6}{c}{NUS-WIDE}\\
\hline 
 &  & \multicolumn{3}{c}{Alex} & \multicolumn{3}{c}{VGG} & \multicolumn{3}{c}{Alex} & \multicolumn{3}{c}{VGG}\\
\cmidrule(lr){3-5}\cmidrule(lr){6-8}\cmidrule(lr){9-11}\cmidrule(lr){12-14}
&  & 16bits& 24bits& 36bits & 16bits& 24bits& 36bits& 16bits& 24bits& 36bits& 16bits& 24bits& 36bits \\
\midrule
\multirow{4}{*}{SDC} 
& Clean & 81.0 & 83.9 & 84.1 & 83.6 & 86.7 & 86.9 & 65.7 & 66.0 & 66.1 & 72.1 & 73.6 & 73.6  \\
& Basic  & 27.5 & 22.1 & 26.7 & 19.9 & 14.8 & 13.0 & 47.7 & 47.2 & 46.8 & 37.0 & 34.7 & 33.9  \\
& APD  & 13.7 & 15.7 & 17.5 & 16 & 10.9 & 9.1 & 31.5 & 29.8 & 28.9  & 30.9 & 29.2 & 28.1 \\
&\textbf{AOD} & \underline{\textbf{12.3}} & \underline{\textbf{14.8}} & \underline{\textbf{17.2}} & \underline{\textbf{13.6}} & \underline{\textbf{9.5}} & \underline{\textbf{8.0}} & \underline{\textbf{28.9}} & \underline{\textbf{28.9}} & \underline{\textbf{28.8}} & \underline{\textbf{28.5}} & \underline{\textbf{26.7}} & \underline{\textbf{25.7}} \\
\midrule
\midrule

\multirow{4}{*}{ADC} 
& Clean & 81.3 & 84.5 & 84.6 & 84.0 & 86.3 & 86.7 & 66.2 & 66.1 & 65.9 & 73.2 & 73.4 & 73.3  \\
& Basic  & 25.3 & 22.1 & 26.7 & 18 & 13.4 & 12.2 & 46.5 & 46.3 & 46.2 & 35.0 & 33.5 & 33.0  \\
& APD  & 10.3 & 14.4 & 17.2 & 13.5 & 9.4 & 8.2 & 30.3 & 28.9 & \textbf{28.3} & 28.6 & 27.5 & 27.0  \\
&\textbf{AOD} & \underline{\textbf{10.2}} & \underline{\textbf{13.2}} & \underline{\textbf{16.6}} & \underline{\textbf{12.0}} & \underline{\textbf{8.5}} & \underline{\textbf{7.4}} & \underline{\textbf{28.2}} & \underline{\textbf{28.1}} & \underline{28.5} & \underline{\textbf{26.3}} & \underline{\textbf{25.0}} & \underline{\textbf{24.6}}  \\
\bottomrule
\end{tabular}
}
\end{center}

\caption{The white-box attack results for a different number of bits of two models on CIFAR-10 and NUS-WIDE. Lower mAP means better attack performance.}
\label{whitebox}
\end{table*}

\subsection{Implementations}
Unless otherwise stated, the targeted deep PQ models are constructed with the most common configuration. First, the base models (e.g. AlexNet, VGG) are pretrained on the evaluated datasets. Afterwards, we extract features from the top layer of the pretrained networks for each of the database images. Then the feature vectors are partitioned into M sub-vectors and K-means clustering is  conducted for each of the M subspace, obtaining the respective codebooks $\{\bf c_{mk}\}_{m=1,k=1}^{M,K}$. 

All the experiments are conducted based on the PyTorch framework. 
 
PQ-AG methods are implemented based on AdverTorch \cite{ding2019advertorch} Framework. For adversarial query generation, we adopt PGD \cite{iclr/KurakinGB17a} to optimize our proposed algorithms. The learning rate is set to 0.01, and each query is trained for 5 iterations. The adversarial purturbations are restricted within 8 by $L_{\inf}$ (pixel values are clipped in [0, 255]).

\subsection{Attack Evaluation}

We evaluate the effectiveness of our proposed PQ-AG algorithm containing two types of attacks, i.e., Type I: \textbf{A}ttak on the \textbf{P}eak of Centroid \textbf{D}istribution (APD)
and Type II: \textbf{A}ttak on the \textbf{O}verall Centroid \textbf{D}istribution (AOD) on the datasets discussed above. The basic attack (denoted as basic) introduced in (\ref{basic}) is also evaluated in the experiments. In total, the experiments are conducted on three well-known network structures (AlexNet, VGG, and ResNet, each with multiple models). The performance of the proposed methods under white-box and-black box settings will be examined in the following experiments, respectively.

\begin{figure}[htb]
\begin{center}
\includegraphics[width=0.95\columnwidth]{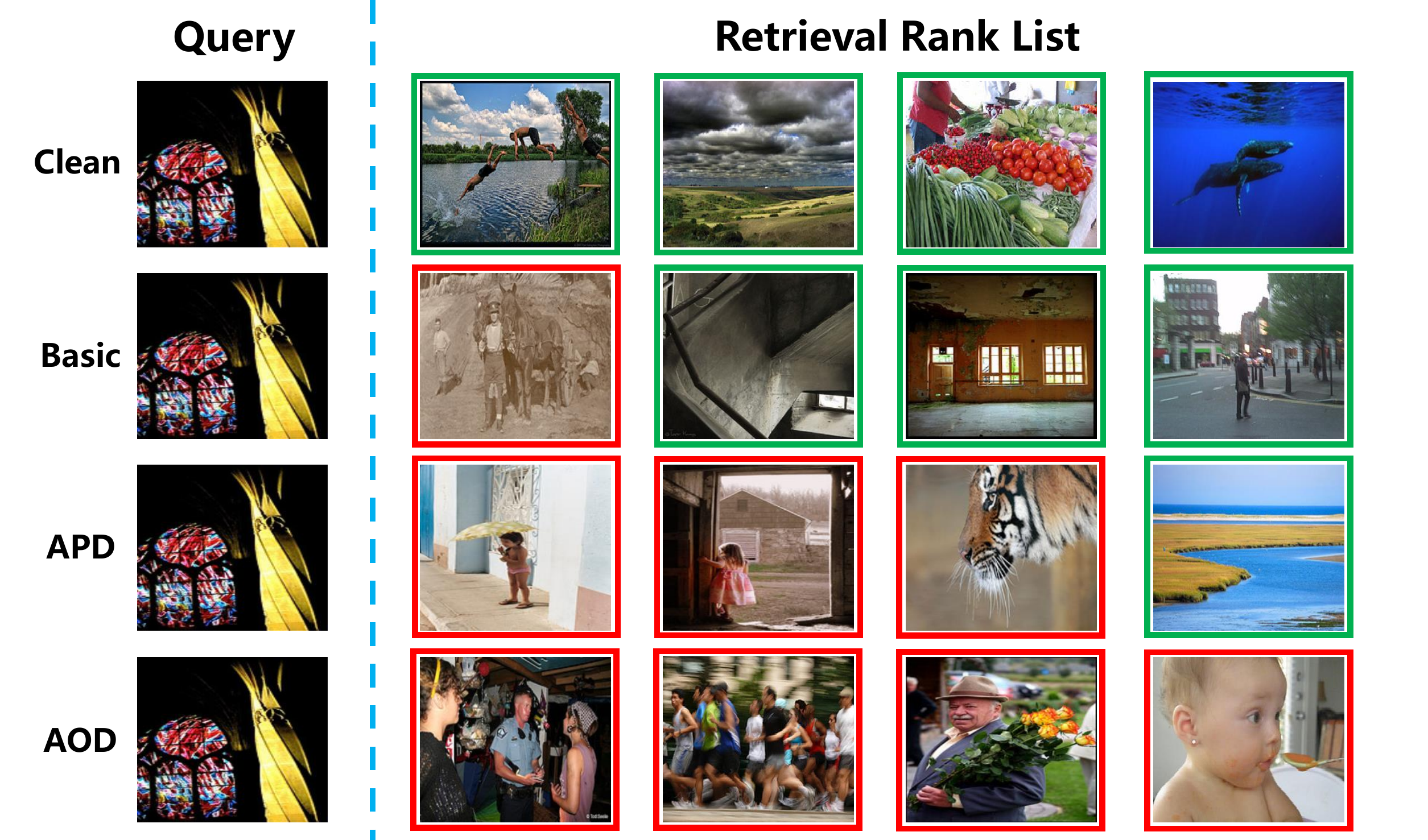}

\end{center}
  \caption{Examples of retrieval results for the respective attack, where irrelevant retrieval results are labeled with red borders.}
\label{demo}
\end{figure}

\noindent \textbf{White-Box Attack}
We first evaluate the white-box effectiveness of the proposed attacks against basic attack across encoding bits. To make a more comprehensive comparison, we use AlexNet and VGG16 in white-box experiments. The mAP for the datasets under various attacks with the encoding bits ranging from 16 bits to 32 bits, are shown in Tab. \ref{whitebox}. Clearly, it can be seen from Tab. \ref{whitebox} that modern product quantization based image retrieval models are extremely vulnerable to adversarial queries. Even under the simplest basic attack, the mAP value drops significantly by 57.7\% on average. Then APD attack consistently outperforms the Basic Attack across models and encoding bits for both  CIFAR-10 and  NUS-WIDE datasets, which reveals that for product quantization systems, manipulating the original quantization results is an effective means of increasing the quantization error, thus result in the  precision  dropping drastically. AOD attack shows its priority over APD attack, and continue to diminish the mAP for most settings. Such an improved result means that, despite its success in image classification attacks, decreasing the probability for the most probable centroids (class in the case of image classification) may not be the best solution to image retrieval attacks. Instead, we should disrupt the overall distribution of the codebook. Some visualization can be found in Fig. \ref{demo}.

\begin{figure*}[htb]
\centering
\smallskip
\begin{subfigure}{.24\linewidth}
  \centering
  \includegraphics[width=\textwidth]{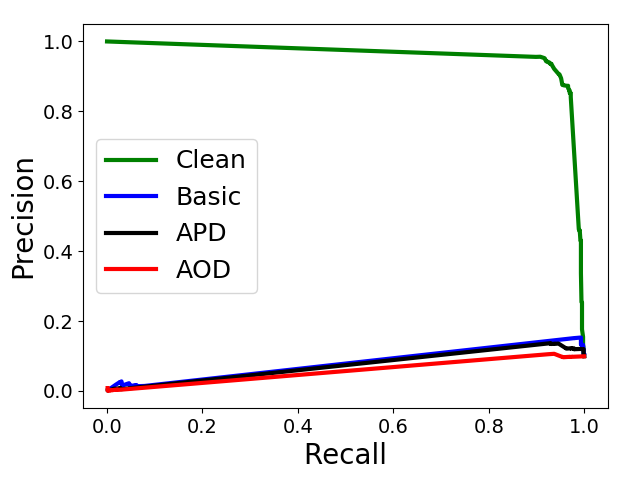}
\caption{ AlexNet on CIFAR-10}
  \label{fig:eh1}
\end{subfigure}
\begin{subfigure}{.24\linewidth}
  \centering
  \includegraphics[width=\textwidth]{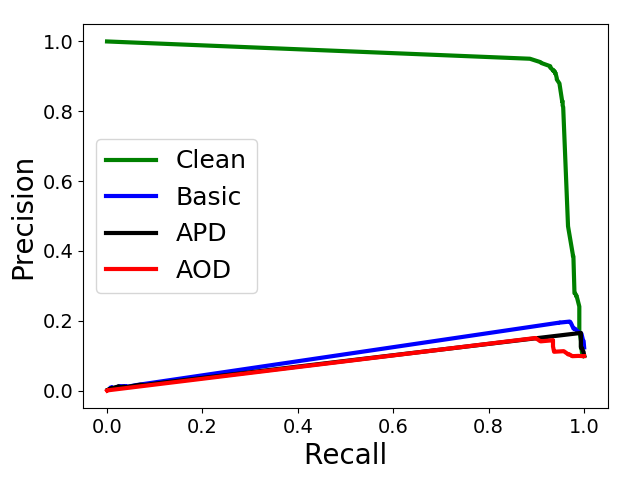}
\caption{VGG16 on CIFAR-10}
  \label{fig:eh2}
\end{subfigure}
\begin{subfigure}{.24\linewidth}
  \centering
  \includegraphics[width=\textwidth]{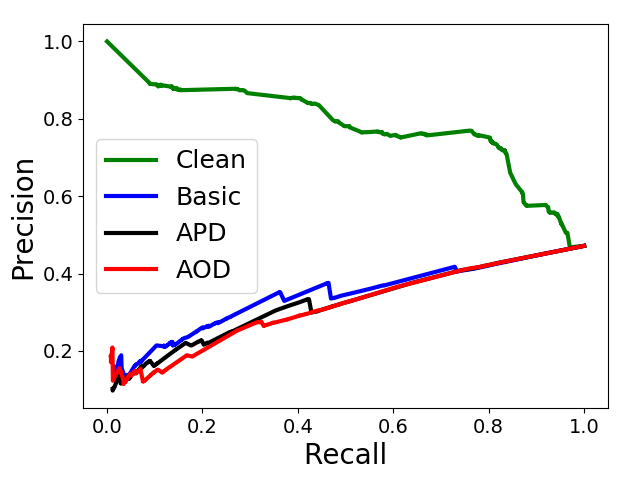}

\caption{AlexNet on NUS-WIDE}
  \label{fig:eh3}
\end{subfigure}
\begin{subfigure}{.24\linewidth}
  \centering
  \includegraphics[width=\textwidth]{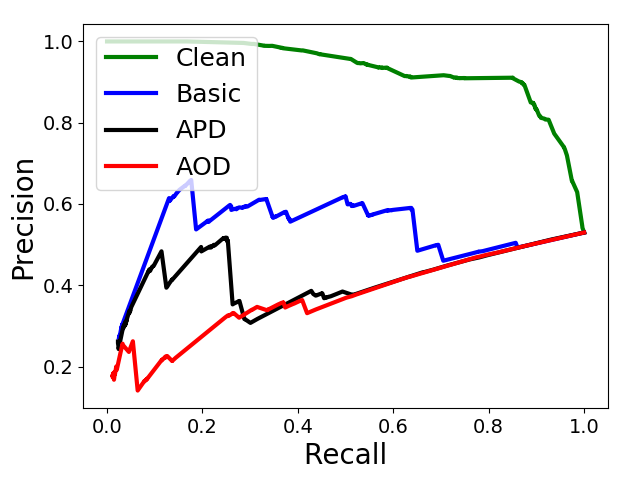}

\caption{VGG16 on NUS-WIDE}
  \label{fig:eh4}
\end{subfigure}
 \caption{PR curves for AlexNet and VGG16 on CIFAR-10 and NUS-WIDE, respectively.}
\label{prcurve}

\end{figure*}

 Precision Recall (PR) curves for the proposed methods on  CIFAR-10 and  NUS-WIDE are shown in Fig. \ref{prcurve}. For the fairness of comparison and better visualization, all the curves are calculated with 16-bits quantization. We can see from Fig. \ref{prcurve} that, different from the original one, PR curves for adversarial examples have a monotone increasing tendency and become furthest away from the original curves for the top retrieval regions, where the recall precision is low. This demonstrates that the proposed attacks completely disrupt the semantic information in the feature space, and effectively pushes the features of adversarial queries away from the originally relevant images in the database.

 We also demonstrate the effectiveness of the proposed attacks on the state-of-the-art DPQN. More specifically, we adopt the method proposed in \cite{eccv/YuYFJ18}, and jointly train the retrieval model with PQ code-book in an end-to-end manner. The attack results against state-of-the-art retrieval models on CIFAR-10 are shown in Tab. \ref{eccv}. As can be seen from Tab. \ref{eccv}, despite the improved performance on clean data, the jointly trained models still suffer from the lack of robustness. In fact, comparing Tab. \ref{eccv} with Tab. \ref{whitebox}, we could find that the jointly trained models are even less robust than the pretrained models against our PQ-AG attacks. These results indicate that robustness should also be included in evaluation when designing DPQNs in the future.

\begin{table}[htb]

\smallskip
\begin{center}

\begin{tabular}{c<{\centering}  c<{\centering} c<{\centering} c<{\centering} c<{\centering}}

\toprule
\multicolumn{2}{c}{}  & 16bits & 24bits & 36bits\\

\midrule

\multirow{4}{*}{Alex} 
& Clean & 87.9 / 87.8 & 89.6 / 89.4 & 89.7 / 89.5  \\

& Basic  & 17.8 / 12.8 & 12.1 / 9.5 & 9.7 / 8.4 \\

& APD  & 15.0 / 9.5 & 9.7 / 6.6 & 7.4 / 5.7 \\

& \textbf{AOD} & 11.5 / 8.8 & 8.2 / 6.5 & 6.5 / 5.8 \\

\bottomrule
\end{tabular}
\end{center}

\caption{White-box attack results for the jointly trained model on CIFAR-10.}
\label{eccv}
\end{table}

\noindent \textbf{Black-Box Attack}
We try to evaluate the Type II PQ-AG attack in the black-box settings from two aspects as follows: 
\begin{itemize}
\item {\bf Bits Transferibility:}  For a given targeted model with various code lengths, the generated adversarial queries can be transferable. 
\item {\bf Model Transferibility:} For various model architectures, the generated adversarial queries can be transferable. 
\end{itemize}

\begin{table}[htb]

\smallskip
\begin{center}

\resizebox{0.95\columnwidth}{!}{
\begin{tabular}{c<{\centering}  c<{\centering} c<{\centering} c<{\centering} c<{\centering}}

\toprule
\multicolumn{2}{c}{}  & 16bits & 24bits & 36bits\\

\midrule

\multirow{4}{*}{Alex} 
& Clean & 81.0 / 81.3 & 83.9 / 84.5 & 84.1 / 84.6  \\

& 16bits  & 12.3 / 10.2 & 13.6 / 13.4 & 17.0 / 15.4 \\

& 24bits  & 15.5 / 14.3 & 14.8 / 13.2 & 16.7 / 15.7 \\

& 36bits & 17.6 / 19.2 & 15.9 / 15.1 & 17.2 / 16.6 \\
\midrule
\midrule

\multirow{4}{*}{VGG} 
& Clean & 83.6 / 84.0 & 86.7 / 86.3 & 86.9 / 86.7  \\

& 16bits  & 13.6 / 12.0 & 9.6 / 8.6 & 8.4 / 7.8 \\

& 24bits  & 13.8 / 13.3 & 9.5 / 8.5 & 8.3 / 7.7 \\

& 36bits & 13.5 / 11.8 & 9.4 / 8.4 & 8.0 / 7.4 \\

\bottomrule
\end{tabular}
}
\end{center}

\caption{Evaluation results for bits transferability.}
\label{crossbit}
\end{table}

The bits transferability of PQ-AG can be found in Tab. \ref{crossbit}. Each row header of Tab. \ref{crossbit} represents the code length of the model for generating the clean and adversarial queries, and each column header represents the code length of the targeted model. The experiments are conducted on CIFAR-10 for both AlexNet and VGG16 models. As can be seen from the table, the mAP values (SDC/ADC) in each column are really close, which verifies the transferability of PQ-AG across bits. This could be attributed to the fact that the centroids in the codebooks across code lengths are essentially cluster centers from the same feature space. Therefore, centroids in a larger code book could be seen as interpolation among the ones in a smaller codebook. As a result, perturbing the centroid distribution for one of the code lengths could lead to the corresponding distributions of all code lengths to be disrupted.

\begin{table}[htb]
\smallskip
\begin{center}
\resizebox{0.95\columnwidth}{!}{
\begin{tabular}{c<{\centering}  c<{\centering} c<{\centering} c<{\centering} c<{\centering} |c<{\centering} c<{\centering} c<{\centering} }
\toprule
\multicolumn{2}{c}{}  & Res18 & Res34 & Res50 & VGG11 & VGG13 & VGG16 \\
\midrule

\multirow{7}{*}{SDC} 
& Clean & 87.7 & 89.2 & 87.1 & 84.3 & 85.0 & 83.6  \\
\cmidrule(lr){2-8}
& Res18 & 9.5 & 36.8 & 46.0 & 66.9 & 67.5 & 64.5\\
& Res34 & 30.8 & 10.8 & 33.8 & 62.5 & 62.2 & 60.0\\
& Res50 & 38.6 & 32.7 & 10.5 & 65.4 & 65.4 & 62.4\\
\cmidrule(lr){2-8}
& VGG11  & 71.0 & 73.0 & 73.5 & 12.6 & 61.6 &64.1\\
& VGG13  & 61.1 & 64.0 & 64.0 & 50.1 & 12.2 &49.7\\
& VGG16 & 60.2 & 58.6 & 61.7 & 53.4 & 49.8 & 13.6\\

\midrule
\midrule

\multirow{7}{*}{ADC} 
& Clean & 87.8 & 89.5 & 87.8  & 84.7 & 85.2 & 84.0\\
\cmidrule(lr){2-8}
& Res18 & 7.7 & 38.5 & 47.2 & 67.9 & 69.0 & 66.0\\
& Res34 & 30.3 & 8.3 & 34.2 & 63.6 & 63.6 & 61.3\\
& Res50 & 39.9 & 34.5 & 8.3 & 66.5 & 67.0 & 64.3\\
\cmidrule(lr){2-8}
& VGG11  & 72.1 & 74.2 & 75.3 & 9.8 & 63.3 &65.8\\
& VGG13  & 63.0 & 65.7 & 66.0 & 52.0 & 9.6 &51.6\\
& VGG16 & 61.3 & 60.8 & 63.3 & 55.2 & 51.6 & 12.0\\

\bottomrule
\end{tabular}
}
\end{center}

\caption{Evaluation results for model transferability.}
\label{crossmodel}
\end{table}

Tab. \ref{crossmodel} demonstrates the transferability of PQ-AG among different CNN models. We conduct the experiments on  CIFAR-10 for ResNet and VGG models. As before, each row header represents the model, form which the adversarial queries are generated, while each column header represents the structure of the targeted model. For both SDC and ADC product quantization retrieval systems, we divide the table into four quarters. The diagonal quarters are designed to evaluate the transferability of PQ-AG among the models from the same families (e.g. VGG11, VGG13, VGG16 are all from the VGG family), ensuring the models share a certain amount of similarity. Therefore, we can see from Tab. \ref{crossmodel} that the mAP drops by a large margin of 68.5\% and 51.6\% respectively for ResNet and VGG models in this two quarters on average. However, adversarial queries in other two quarters are generated against one kind of models and designed to fool models from completely different families. Therefore, the task is obviously much more challenging. Even so, PQ-AG still manages to decrease the mAP by 24.9\% on average in all these cases.

\section{Conclusion}
A novel PQ-AG algorithm is developed for generating adversarial queries to fool the deep product quantization retreieval systems. With the estimation of the centroid distribution, we propose a differentiable loss for back propagation to generate adversarial queries.
Extensive  experiments are conducted on two public datasets based on multiple network structures, demonstrating that PQ-AG can effectively generate adversarial queries that induce the PQ based retrieval models into retrieving semantically irrelevant results. In the meantime, the transferability of the generated queries is also evaluated under a variety of settings, revealing that PQ-AG also works in black-box setting. 

\section{Acknowledgement}
This work is supported in part by the National Key Research and Development Program of China under Grant 2018YFB1800204, the National Natural Science Foundation of China under Grant 61771273, the China Postdoctoral Science Foundation under Grant 2019M660645, the R\&D Program of Shenzhen under Grant JCYJ20180508152204044, and the research fund of PCL Future Regional Network Facilities for Large-scale Experiments and Applications (PCL2018KP001).

\bibliographystyle{aaai}
\bibliography{aaai.bib}

\begin{thebibliography}{}

\bibitem[\protect\citeauthoryear{Babenko and Lempitsky}{2014}]{AQ}
Babenko, A., and Lempitsky, V.~S.
\newblock 2014.
\newblock Additive quantization for extreme vector compression.
\newblock In {\em CVPR}.

\bibitem[\protect\citeauthoryear{Bhagoji \bgroup et al\mbox.\egroup
  }{2018}]{eccv/BhagojiHLS18}
Bhagoji, A.~N.; He, W.; Li, B.; and Song, D.
\newblock 2018.
\newblock Practical black-box attacks on deep neural networks using efficient
  query mechanisms.
\newblock In {\em ECCV}.

\bibitem[\protect\citeauthoryear{Cao \bgroup et al\mbox.\egroup
  }{2016}]{CaoL0ZW16}
Cao, Y.; Long, M.; Wang, J.; Zhu, H.; and Wen, Q.
\newblock 2016.
\newblock Deep quantization network for efficient image retrieval.
\newblock In {\em AAAI}.

\bibitem[\protect\citeauthoryear{Carlini and Wagner}{2017}]{sp/Carlini017}
Carlini, N., and Wagner, D.~A.
\newblock 2017.
\newblock Towards evaluating the robustness of neural networks.
\newblock In {\em SP}.

\bibitem[\protect\citeauthoryear{Chen \bgroup et al\mbox.\egroup
  }{2017}]{corr/abs-1712-02051}
Chen, H.; Zhang, H.; Chen, P.; Yi, J.; and Hsieh, C.
\newblock 2017.
\newblock Show-and-fool: Crafting adversarial examples for neural image
  captioning.
\newblock {\em arXiv: 1712.02051}.

\bibitem[\protect\citeauthoryear{Chua \bgroup et al\mbox.\egroup
  }{2009}]{civr/ChuaTHLLZ09}
Chua, T.; Tang, J.; Hong, R.; Li, H.; Luo, Z.; and Zheng, Y.
\newblock 2009.
\newblock {NUS-WIDE:} a real-world web image database from national university
  of singapore.
\newblock In {\em CIVR}.

\bibitem[\protect\citeauthoryear{Datar \bgroup et al\mbox.\egroup
  }{2004}]{datar2004locality}
Datar, M.; Immorlica, N.; Indyk, P.; and Mirrokni, V.~S.
\newblock 2004.
\newblock Locality-sensitive hashing scheme based on p-stable distributions.
\newblock In {\em Proceedings of the twentieth annual symposium on
  Computational geometry}.
\newblock ACM.

\bibitem[\protect\citeauthoryear{Ding, Wang, and
  Jin}{2019}]{ding2019advertorch}
Ding, G.~W.; Wang, L.; and Jin, X.
\newblock 2019.
\newblock {AdverTorch} v0.1: An adversarial robustness toolbox based on
  pytorch.
\newblock {\em arXiv preprint arXiv:1902.07623}.

\bibitem[\protect\citeauthoryear{G, T, and P}{2006}]{MITbook}
G, S.; T, D.; and P, I.
\newblock 2006.
\newblock Nearest-neighbor methods in learning and vision: Theory and practice.
\newblock {\em ch.3, MIT Press}.

\bibitem[\protect\citeauthoryear{Ge \bgroup et al\mbox.\egroup }{2013}]{OPQ}
Ge, T.; He, K.; Ke, Q.; and Sun, J.
\newblock 2013.
\newblock Optimized product quantization for approximate nearest neighbor
  search.
\newblock In {\em CVPR}.

\bibitem[\protect\citeauthoryear{Gionis \bgroup et al\mbox.\egroup
  }{1999}]{gionis1999similarity}
Gionis, A.; Indyk, P.; Motwani, R.; et~al.
\newblock 1999.
\newblock Similarity search in high dimensions via hashing.
\newblock In {\em Vldb}.

\bibitem[\protect\citeauthoryear{Goodfellow, Shlens, and
  Szegedy}{2015}]{corr/GoodfellowSS14}
Goodfellow, I.~J.; Shlens, J.; and Szegedy, C.
\newblock 2015.
\newblock Explaining and harnessing adversarial examples.
\newblock In {\em ICLR}.

\bibitem[\protect\citeauthoryear{J{\'{e}}gou, Douze, and Schmid}{2013}]{pq}
J{\'{e}}gou, H.; Douze, M.; and Schmid, C.
\newblock 2013.
\newblock Product quantization for nearest neighbor search.
\newblock {\em TPAMI}.

\bibitem[\protect\citeauthoryear{Kalantidis and
  Avrithis}{2014}]{kalantidis2014locally}
Kalantidis, Y., and Avrithis, Y.
\newblock 2014.
\newblock Locally optimized product quantization for approximate nearest
  neighbor search.
\newblock In {\em CVPR}.

\bibitem[\protect\citeauthoryear{Klein and Wolf}{2019}]{Klein_2019_CVPR}
Klein, B., and Wolf, L.
\newblock 2019.
\newblock End-to-end supervised product quantization for image search and
  retrieval.
\newblock In {\em CVPR}.

\bibitem[\protect\citeauthoryear{Krizhevsky and
  Hinton}{2009}]{krizhevsky2009learning}
Krizhevsky, A., and Hinton, G.
\newblock 2009.
\newblock Learning multiple layers of features from tiny images.

\bibitem[\protect\citeauthoryear{Kurakin, Goodfellow, and
  Bengio}{2017}]{iclr/KurakinGB17a}
Kurakin, A.; Goodfellow, I.~J.; and Bengio, S.
\newblock 2017.
\newblock Adversarial examples in the physical world.
\newblock In {\em ICLR}.

\bibitem[\protect\citeauthoryear{Li \bgroup et al\mbox.\egroup
  }{2017}]{li2017distribution}
Li, L.; Hu, Q.; Han, Y.; and Li, X.
\newblock 2017.
\newblock Distribution sensitive product quantization.
\newblock {\em TCSVT}.

\bibitem[\protect\citeauthoryear{Liu \bgroup et al\mbox.\egroup
  }{2019}]{liu2019deep}
Liu, B.; Cao, Y.; Long, M.; Wang, J.; and Wang, J.
\newblock 2019.
\newblock Deep triplet quantization.
\newblock {\em arXiv preprint arXiv:1902.00153}.

\bibitem[\protect\citeauthoryear{Madry \bgroup et al\mbox.\egroup
  }{2018}]{iclr/MadryMSTV18}
Madry, A.; Makelov, A.; Schmidt, L.; Tsipras, D.; and Vladu, A.
\newblock 2018.
\newblock Towards deep learning models resistant to adversarial attacks.
\newblock In {\em ICLR}.

\bibitem[\protect\citeauthoryear{Moosavi{-}Dezfooli \bgroup et al\mbox.\egroup
  }{2017}]{cvpr/Moosavi-Dezfooli17}
Moosavi{-}Dezfooli, S.; Fawzi, A.; Fawzi, O.; and Frossard, P.
\newblock 2017.
\newblock Universal adversarial perturbations.
\newblock In {\em CVPR}.

\bibitem[\protect\citeauthoryear{Muja and Lowe}{2009}]{muja2009fast}
Muja, M., and Lowe, D.~G.
\newblock 2009.
\newblock Fast approximate nearest neighbors with automatic algorithm
  configuration.
\newblock {\em VISAPP (1)}.

\bibitem[\protect\citeauthoryear{Szegedy \bgroup et al\mbox.\egroup
  }{2014}]{corr/SzegedyZSBEGF13}
Szegedy, C.; Zaremba, W.; Sutskever, I.; Bruna, J.; Erhan, D.; Goodfellow,
  I.~J.; and Fergus, R.
\newblock 2014.
\newblock Intriguing properties of neural networks.
\newblock In {\em ICLR}.

\bibitem[\protect\citeauthoryear{Wang and Zhang}{2019}]{CQ}
Wang, J., and Zhang, T.
\newblock 2019.
\newblock Composite quantization.
\newblock {\em TPAMI}.

\bibitem[\protect\citeauthoryear{Wang \bgroup et al\mbox.\egroup
  }{2014}]{wang2014optimized}
Wang, J.; Wang, J.; Song, J.; Xu, X.-S.; Shen, H.~T.; and Li, S.
\newblock 2014.
\newblock Optimized cartesian k-means.
\newblock {\em TKDE}.

\bibitem[\protect\citeauthoryear{Wang \bgroup et al\mbox.\egroup }{2016}]{SQ}
Wang, X.; Zhang, T.; Qi, G.; Tang, J.; and Wang, J.
\newblock 2016.
\newblock Supervised quantization for similarity search.
\newblock In {\em CVPR}.

\bibitem[\protect\citeauthoryear{Wu \bgroup et al\mbox.\egroup
  }{2017}]{wu2017multiscale}
Wu, X.; Guo, R.; Suresh, A.~T.; Kumar, S.; Holtmann-Rice, D.~N.; Simcha, D.;
  and Yu, F.
\newblock 2017.
\newblock Multiscale quantization for fast similarity search.
\newblock In {\em NeurIPS}.

\bibitem[\protect\citeauthoryear{Xie \bgroup et al\mbox.\egroup
  }{2017}]{iccv/XieWZZXY17}
Xie, C.; Wang, J.; Zhang, Z.; Zhou, Y.; Xie, L.; and Yuille, A.~L.
\newblock 2017.
\newblock Adversarial examples for semantic segmentation and object detection.
\newblock In {\em ICCV}.

\bibitem[\protect\citeauthoryear{Yang \bgroup et al\mbox.\egroup
  }{2018}]{Yang2018Adversarial}
Yang, E.; Liu, T.; Deng, C.; and Tao, D.
\newblock 2018.
\newblock Adversarial examples for hamming space search.
\newblock {\em IEEE transactions on cybernetics}.

\bibitem[\protect\citeauthoryear{Yu \bgroup et al\mbox.\egroup
  }{2017}]{yu2017bilinear}
Yu, L.; Huang, Z.; Shen, F.; Song, J.; Shen, H.~T.; and Zhou, X.
\newblock 2017.
\newblock Bilinear optimized product quantization for scalable visual content
  analysis.
\newblock {\em TIP}.

\bibitem[\protect\citeauthoryear{Yu \bgroup et al\mbox.\egroup
  }{2018}]{eccv/YuYFJ18}
Yu, T.; Yuan, J.; Fang, C.; and Jin, H.
\newblock 2018.
\newblock Product quantization network for fast image retrieval.
\newblock In {\em ECCV}.

\bibitem[\protect\citeauthoryear{Zhao \bgroup et al\mbox.\egroup
  }{2019}]{Zhao2019Gan}
Zhao, G.; Zhang, M.; Liu, J.; and Wen, J.-R.
\newblock 2019.
\newblock Unsupervised adversarial attacks on deep feature-based retrieval with
  {GAN}.
\newblock {\em arXiv: 1907.05793}.

\end{thebibliography}

\end{document}